\documentclass[10pt,twocolumn,letterpaper]{article}

\usepackage[pagenumbers]{iccv} 

\definecolor{iccvblue}{rgb}{0.21,0.49,0.74}
\usepackage[pagebackref,breaklinks,colorlinks,allcolors=iccvblue]{hyperref}

\usepackage{multirow}
\usepackage{float}
\usepackage{stfloats}
\usepackage{placeins}
\usepackage{adjustbox}

\def\paperID{7995} 
\def\confName{ICCV}
\def\confYear{2025}

\title{FROSS: Faster-than-Real-Time Online \\3D Semantic Scene Graph Generation from RGB-D Images}

\author{
Hao-Yu Hou$^{1,3}$ \and Chun-Yi Lee$^2$ \and Motoharu Sonogashira$^3$ \and Yasutomo Kawanishi$^3$ \and
\\$^1$National Tsing Hua University \and \\$^2$National Taiwan University \and \\$^3$RIKEN
}

\begin{document}

\maketitle

\begin{abstract}
The ability to abstract complex 3D environments into simplified and structured representations is crucial across various domains. 3D semantic scene graphs (SSGs) achieve this by representing objects as nodes and their interrelationships as edges, facilitating high-level scene understanding. Existing methods for 3D SSG generation, however, face significant challenges, including high computational demands and non-incremental processing that hinder their suitability for real-time open-world applications. To address this issue, we propose FROSS (\textbf{F}aster-than-\textbf{R}eal-Time \textbf{O}nline 3D \textbf{S}emantic \textbf{S}cene Graph Generation), an innovative approach for online and faster-than-real-time 3D SSG generation that leverages the direct lifting of 2D scene graphs to 3D space and represents objects as 3D Gaussian distributions. This framework eliminates the dependency on precise and computationally-intensive point cloud processing. Furthermore, we extend the Replica dataset with inter-object relationship annotations, creating the ReplicaSSG dataset for comprehensive evaluation of FROSS. The experimental results from evaluations on ReplicaSSG and 3DSSG datasets show that FROSS can achieve superior performance while operating significantly faster than prior 3D SSG generation methods. Our implementation and dataset are publicly available at \mbox{\url{https://github.com/Howardkhh/FROSS}}.

\vspace{-1em}
\end{abstract}

\section{Introduction}
\label{sec:intro}

\begin{figure}[t]
    \centering
    \includegraphics[width=0.475\textwidth]{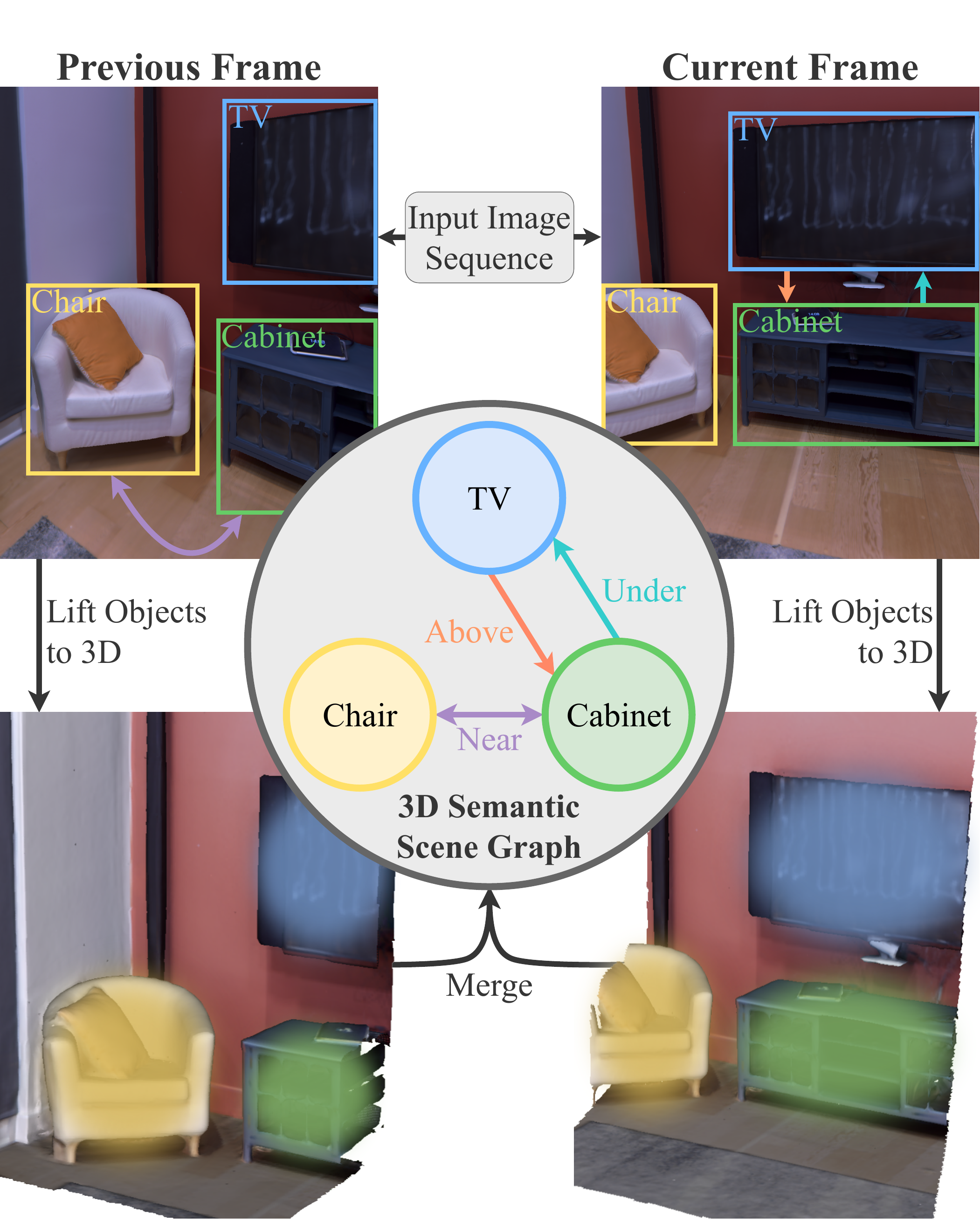}
    \caption{We introduce FROSS, an online real-time 3D semantic scene graph generation method that leverages and integrates 2D scene graphs. FROSS represents objects as 3D Gaussian distributions and operates without requiring 3D reconstruction.}
    \vspace{-0.5em}
    \label{fig:teaser}
\end{figure}

The abstraction of complex 3D environments into simplified representations in real time serves as a crucial capability across multiple application domains. The incremental generation of such representations in real time holds particular significance for fields ranging from robotics and computer vision to augmented reality (AR), where systems must perceive and analyze the environment instantaneously. 3D scene graphs (SGs)~\cite{gay2019visual, armeni20193d} accomplish this abstraction by representing physical objects as nodes and their interrelationships as edges within a structured framework. 3D semantic scene graphs (SSGs)~\cite{wald2020learning} extend this representation with an emphasis on semantic relationships, such as positional configurations between objects (e.g., a cup on a table) or functional interactions (e.g., a person seated in a chair), which enables high-level scene comprehension, reasoning, and interaction. Real-world applications, however, present open-world challenges where environments often exceed known spatial boundaries and contain previously unseen spaces~\cite{shah2021ving}. For instance, a room may connect to multiple adjacent spaces or corridors, which necessitates continuous adaptation and incremental scene understanding. This construction process must maintain high efficiency to generate structured environmental representations without creating performance bottlenecks. Meeting the processing demands requires the generation of 3D SSG to operate at \textit{faster-than-real-time} speeds, with all computations completed between successive image frames from a high-speed sensor stream.

Executing all processes at faster-than-real-time rates confers multiple advantages: (1) prevention of bottlenecks in the 3D SSG generation framework, (2) facilitation of smooth operation on resource-constrained hardware such as edge computing devices, and (3) preservation of computational resources for additional system components. Contemporary 3D SSG generation approaches predominantly focus on offline construction, which requires access to the complete scene data, either instance-segmented point clouds~\cite{wald2020learning, zhang2021exploiting}, a complete image sequence~\cite{gay2019visual}, or both~\cite{wang2023vl}. This assumption of full scene availability conflicts with incremental environmental exploration. Although several studies~\cite{wu2021scenegraphfusion, wu2023incremental} have developed methods for incremental point cloud reconstruction and segmentation to achieve online and real-time 3D SSG generation, their aggregate system latency remains substantial, even with the utilization of multiple CPU cores. This constraint stems mainly from the substantial computational demands of environmental mapping and point cloud processing, restricting their practical application in real-time systems with limited computational power or other concurrent processes. Given these limitations, the aim of this study is to develop a method for faster-than-real-time online SSG generation.

A fundamental insight for generating 3D SSGs emerges from the observation that precise object pose and shape information is not essential. Given that 3D SSGs provide a high-level semantic understanding of the environment, their functional implications remain valid with approximate object locations. For instance, in robotic applications, high-level planning can be successfully executed using relative spatial relationships~\cite{fox2003pddl2, hoffmann2001ff, tan2023knowledge}. Similarly, in 3D scene synthesis and room layout generation~\cite{ccelen2024design, feng2024layoutgpt, zheng2024editroom, gao2024graphdreamer, zhai2023commonscenes, zhai2024echoscene}, 3D SSGs serve as foundational structures from which more detailed content can be generated. These application scenarios eliminate the need for precise environmental mapping techniques such as Simultaneous Localization and Mapping (SLAM)~\cite{campos2021orb} and additional point cloud processing, both of which contribute significant computational overhead.

This observation motivates an alternative approach for online and real-time 3D SSG generation: inferring 2D SGs from images and subsequently lifting them into 3D space, thus circumventing the computational burden of environmental mapping and point cloud processing. To the best of our knowledge, the current literature has not adequately investigated this research direction except~\cite{kim20193}. Such unexplored territory presents significant challenges, including the handling of incomplete 2D object observations, the achievement of real-time SG generation, and the effective integration of multi-view information into complete 3D SSGs without object redundancy. These challenges, therefore, provide promising avenues for further innovative research contributions.

In light of these considerations, we introduce FROSS (\textbf{F}aster-than-\textbf{R}eal-time \textbf{O}nline 3D \textbf{S}emantic \textbf{S}cene Graph Generation), a methodology that constructs 3D SSGs by lifting 2D SGs to 3D space and integrating them into a comprehensive representation. To achieve real-time processing without requiring precise object localization, FROSS initiates the process by constructing 2D SGs from detected objects in images. A key innovation of FROSS lies in its utilization of Gaussian distributions to model objects in 3D space, which facilitates the integration of objects observed across multiple images into a coherent 3D SSG, as illustrated in Figure~\ref{fig:teaser}. Specifically, FROSS represents detected objects through Gaussian distributions, lifts these distributions from 2D SGs into 3D space, and merges them according to their predicted classes and statistical distances. We validate FROSS on the 3DSSG dataset~\cite{wald2020learning} and conduct additional evaluations on ReplicaSSG, an extended version of the Replica dataset~\cite{replica19arxiv} that we have enhanced with object relationship annotations. ReplicaSSG incorporates object classes defined in Visual Genome~\cite{krishna2017visual}, a widely-adopted 2D SG dataset. The main contributions of the paper can be summarized as follows:
\begin{itemize}
    \item We introduce FROSS, an innovative methodology for online real-time generation of 3D SSGs. FROSS demonstrates superior performance and significantly faster processing speeds compared to existing baseline methods.
    \item We propose a new merging algorithm based on Gaussian distributions for 3D object integration. This algorithm effectively prevents the proliferation of duplicate or incorrectly detected objects as the input image sequence expands, thereby maintaining computational efficiency.
    \item We develop ReplicaSSG, an extension of the Replica dataset with 3D SSG annotations. Comprehensive evaluations of FROSS on both the 3DSSG and ReplicaSSG datasets demonstrate FROSS's effectiveness and computational efficiency over diverse environmental scenarios.
\end{itemize}

\section{Related Work}
\label{sec:related}

SGs~\cite{johnson2015image, xu2017scene, zellers2018neural, im2024egtr} constitute a fundamental data structure for object and relationship representation within images following their introduction by~\cite{johnson2015image}. This representation has demonstrated substantial utility across various scene understanding tasks~\cite{qian2022scene, gao2018image, li2019relation, yang2019auto, kim2019dense, johnson2018image}. The concept later evolved to encompass 3D spaces, where objects may incorporate attributes such as class labels, visual features, shape characteristics, and affordances. Research in 3D SGs developed into two branches of representation: SSGs introduced in ~\cite{gay2019visual, kim20193, wald2020learning} and hierarchical scene graphs (HSGs) proposed in~\cite{armeni20193d, hughes2022hydra, rosinol2021kimera}. Both frameworks represent 3D objects as nodes with pairwise relationships defined through edges. However, these approaches differ fundamentally in their relationship definitions. SSGs concentrate on semantic relationships between objects, while HSGs organize objects into hierarchical layers that include buildings, rooms, and constituent objects, where edges connect only nodes in identical or adjacent layers. Compared to HSGs, SSGs enable higher-level reasoning capabilities through rich semantic relationship representations. The extension beyond spatial relationships allows SSGs to be applied to broader domains such as robotics, 3D computer vision, and AR.

\subsection{Offline 3D Semantic Scene Graph Generation} 
The initial development of 3D SSG generation focused primarily on offline settings that required either complete instance-segmented point clouds~\cite{wald2020learning, zhang2021exploiting, wang2023vl, zhang2021knowledge, liu2022explore, wald2022learning} or image sequences~\cite{gay2019visual, wang2023vl} as inputs. Such requirements limited their applicability to scenarios with incremental data collection, as they necessitated complete data reprocessing. Despite these limitations, these studies established foundations for 3D SSG generation. The research in~\cite{gay2019visual} pioneered 3D SSG generation by predicting 3D object ellipsoids from image collections and utilizing their geometry and appearance features through a recurrent neural network to extract inter-object relationships. The study in~\cite{wald2020learning} advanced this field through the introduction of the 3DSSG dataset, which enhanced the 3RScan dataset~\cite{Wald2019RIO} with SSG annotations. This dataset represents the first large-scale collection with dense annotations and has become the primary benchmark for evaluating 3D SSG generation methods. The authors of~\cite{wald2020learning} further contributed a generation methodology by utilizing instance-segmented point clouds as input and leveraged PointNet~\cite{qi2017pointnet} for feature extraction as well as a graph convolutional network~\cite{kipf2017semi} for label prediction.

\subsection{Online 3D Semantic Scene Graph Generation} 
Several previous studies have explored 3D SG generation in online settings, with some achieving real-time performance. The research in~\cite{kim20193} introduced a framework that predicts local 3D SSGs from RGB-D images and integrates them incrementally into a global representation. However, their computational requirements render real-time execution infeasible. SceneGraphFusion~\cite{wu2021scenegraphfusion} and the study in~\cite{wu2023incremental} developed methods for online point cloud reconstruction and segmentation, followed by feature extraction for local SSG prediction and global fusion. These frameworks can achieve real-time performance through multi-threaded execution of parallel components. In contrast to these existing methods, our proposed FROSS framework avoids precise object localization by approximating objects with Gaussian distributions. This approximation approach enables faster-than-real-time online 3D SSG generation, allowing the end-to-end processing latency to be below the sensor data acquisition period while maintaining single-threaded execution.

\def\meanthreed#1{\mu^{\text{3D}}_{#1}}
\def\covthreed#1{\Sigma^{\text{3D}}_{#1}}
\def\covthreedprime#1{\Sigma^{\text{3D}\prime}_{#1}}
\def\covthreedpp#1{\Sigma^{\text{3D}\prime\prime}_{#1}}
\def\meantwod#1{\mu^{\text{2D}}_{#1}}
\def\covtwod#1{\Sigma^{\text{2D}}_{#1}}

\begin{figure*}[t]
    \centering
    \includegraphics[width=0.95\textwidth]{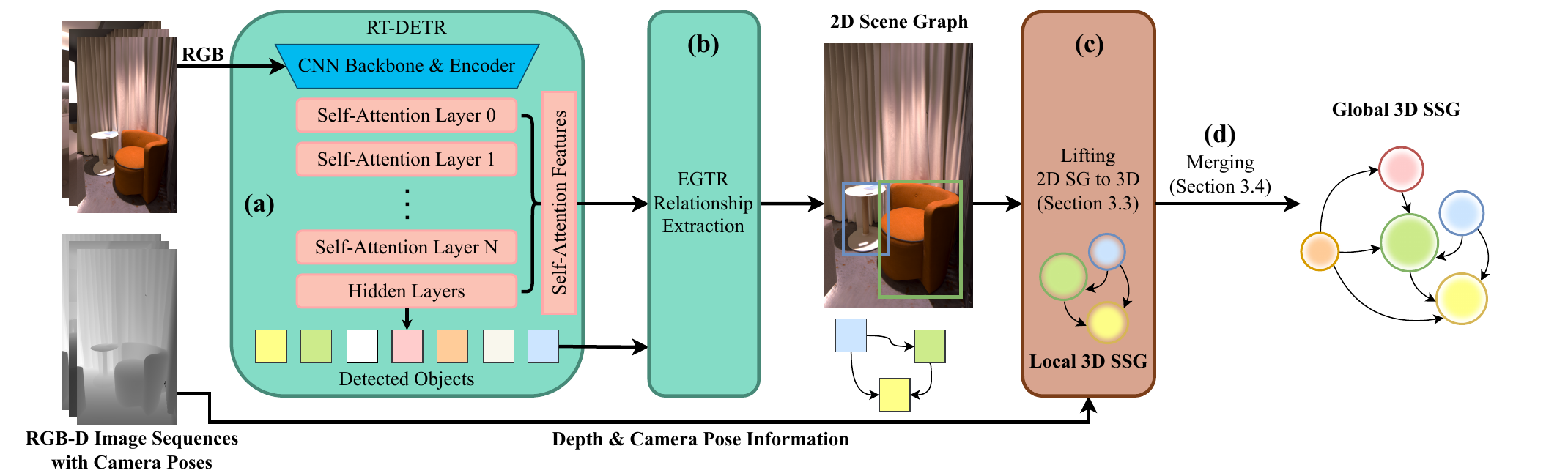}
    \caption{
        An overview of the FROSS framework: (a) The process initiates with object detection via RT-DETR~\cite{zhao2024detrs} from an RGB-D image and its associated camera pose. (b) The EGTR model~\cite{im2024egtr} extracts inter-object relationships through utilization of preserved self-attention features from RT-DETR. (c) 2D Gaussian distributions derived from detected bounding boxes undergo spatial transformation into 3D space for local 3D SSG construction (Section~\ref{sec:methodology:lifting}). (d) The resultant local 3D SSG is subsequently incorporated into the global 3D SSG through our proposed merging algorithm (Section~\ref{sec:methodology:merging}).
    }
    \vspace{-1em}
    \label{fig:overview}
\end{figure*}

\section{Methodology}
\label{sec:methodology}

\subsection{Problem Definition}
\label{sec:methodology:problem}
The primary problem concerned in this study is the online generation of 3D SSGs for environments where the complete scene structure remains unknown a priori. Given a sequence of input images, the primary objective is to construct a 3D SSG $\mathcal{G}$ of the target environment. The graph $\mathcal{G}$ consists of a set of nodes $\mathcal{V}$ and their corresponding directed edges $\mathcal{E}$. Each node $v_i = (c_i, \meanthreed{i}, \covthreed{i}) \in \mathcal{V}$ is indexed by $i \in \{1,\dots,N\}$, representing a unique object in the scene. Every object has a class label $c_i \in \mathcal{C}$, where $\mathcal{C}$ is the set of object categories. While different approaches may incorporate various node attributes, our formulation represents each object as a 3D Gaussian distribution parameterized by mean vector $\meanthreed{i} \in \mathbb{R}^3$ and covariance matrix $\covthreed{i} \in \mathbb{R}^{3\times3}$. The edges $e_{i\rightarrow j} \in \mathcal{E}$ connect node $v_i$ to node $v_j$ with relationship $r_{i\rightarrow j} \in \mathcal{R}$, where $\mathcal{R}$ represents the set of predefined relationship types. Furthermore, the algorithm must be able to maintain online processing capabilities to update $\mathcal{G}$ continuously whenever new image data becomes available.

\subsection{Overview of Framework}
\label{sec:methodology:overview}
Figure~\ref{fig:overview} presents an overview of the proposed FROSS framework, which processes image sequences with corresponding depth information and camera transformations as input. The framework begins with the EGTR~\cite{im2024egtr} 2D SG generation model, with the object detection backbone model replaced with RT-DETR~\cite{zhao2024detrs, lv2024rtdetrv2improvedbaselinebagoffreebies}. This RT-DETR object detector is a state-of-the-art real-time detection model, which preserves intermediate self-attention features for subsequent relationship extraction. FROSS then converts object bounding boxes to 2D Gaussian distributions and projects them into 3D space to generate local SSGs, as elaborated in Section~\ref{sec:methodology:lifting}. These Gaussian distributions serve as efficient approximations of both object positions and their spatial distributions for rapid processing. These local graphs are then integrated into a global SSG through the proposed merging algorithm that is described in detail in Section~\ref{sec:methodology:merging}.

\subsection{Lifting 2D SG to 3D}
\label{sec:methodology:lifting}
The 2D SGs generated from RT-DETR~\cite{zhao2024detrs, lv2024rtdetrv2improvedbaselinebagoffreebies} and EGTR~\cite{im2024egtr} enable FROSS to construct local 3D SSGs $\mathcal{G_\text{local}}=(\mathcal{V_\text{local}},\mathcal{E_\text{local}})$ through the integration of depth information and camera transformation. This process involves two key steps: (1) the conversion of object bounding boxes to 2D Gaussians, and (2) their subsequent back-projection into 3D space while preserving inter-object relationships.

\noindent \textbf{2D Gaussian Representation of Bounding Boxes.} 
Although several methods exist for lifting 2D bounding boxes to 3D space, including point cloud back-projection and joint optimization of 3D ellipsoids~\cite{gay2019visual}, the Gaussian distribution approach emerges as the most computationally efficient solution among these alternatives.
The representation of objects via Gaussian distributions requires the specification of mean $\meantwod{i}$ and covariance matrix $\covtwod{i}$ of the 2D object node $v_i^{\text{2D}}$, which would correspond to a 3D node $v_i \in \mathcal{V}_{\text{local}}$.
Based on the formulation in~\cite{10382963}, FROSS models bounding boxes as 2D uniform distributions. The mean values correspond to the bounding box centers, while the covariance matrices maintain consistency with the uniform distributions. The covariance matrix of $v_i^{\text{2D}}$ is formulated as:
\begin{equation}
    \covtwod{i} = \frac{1}{12}\begin{bmatrix}
                                W_i^2 & 0\\
                                0 & H_i^2
                                \end{bmatrix},
\end{equation}
where $W_i$ and $H_i$ are 2D bounding box's width and height.

\noindent \textbf{3D Back-Projection of Gaussian Distributions.} Given the 2D Gaussians, their corresponding depths at the Gaussian centroids, and camera transformation matrices, the mean positions $\meanthreed{i}$ and covariances $\covthreed{i}$ of the Gaussians in the 3D space can be approximated. The 3D mean $\meanthreed{i}$ can be derived from the 2D mean $\meantwod{i} = (p_x, p_y)^\top$ and its corresponding depth value $z$ through the following equation:
\begin{equation}
    \meanthreed{i} = R \cdot K^{-1} \cdot (p_x, p_y, 1)^\top \cdot z + \mathbf{t},
\end{equation}
where $K \in \mathbb{R}^{3\times3}$ represents the camera intrinsic matrix, and $R \in \mathbb{R}^{3\times3}$ and $\mathbf{t} \in \mathbb{R}^{3}$ denote the rotation matrix and translation vector of the current camera, respectively. 
The approximation of the 3D covariance matrix utilizes the Jacobian $J$ of the local affine approximation of the projective transformation~\cite{zwicker2002ewa}, which is defined as the following:
\begin{equation}
    J=\begin{bmatrix}
        \frac{f_x}{z} & 0 & -\frac{f_xp_x}{z^2} \\
        0 & \frac{f_y}{z} & -\frac{f_yp_y}{z^2}
    \end{bmatrix},
\end{equation}
where $f_x$ and $f_y$ represent the focal lengths in $x$ and $y$ dimensions.
In the original literature of~\cite{zwicker2002ewa}, the Jacobian $J$ is utilized to project 3D covariance matrices onto the 2D image plane. However, our approach inverts this process by back-projecting the 2D covariance matrix $\covtwod{i}$ into 3D space as $\bar{\Sigma}^{\text{3D}}_{i}$ via the pseudo-inverse $J^+$ of the Jacobian:
\begin{equation}    \bar{\Sigma}^{\text{3D}}_{i}=RJ^+\covtwod{i}J^{+\top}R^\top.
\end{equation}
The naive definition alone is insufficient since the original equation is inherently designed for projection from 3D to 2D space. Simply inverting the equation results in a 3D Gaussian that lacks variance information along the depth axis, as the 2D covariance matrix does not capture depth-related variance. This limitation necessitates the specification of variance in the depth dimension. We hypothesize that this variance approximates the average variance of the other dimensions.
The specification has to be applied prior to the rotation matrix $R$, leading to the following equations:
\begin{align}
    \covthreedpp{i}&=J^+\covtwod{i}J^{+\top}, \\
    \covthreedprime{i}&=\covthreedpp{i} + \begin{bmatrix} 
                                        0 & 0 & 0 \\ 
                                        0 & 0 & 0 \\
                                        0 & 0 & \frac{(\covthreedpp{i})_{1, 1} + (\covthreedpp{i})_{2, 2}}{2}
                                        \end{bmatrix}.
\end{align}
In this formulation, $(\covthreedpp{i})_{m, n} \in \mathbb{R}$ represents the element at position $(m,n)$ of matrix $\covthreedpp{i}$. Finally, the covariance matrix $\covthreedprime{i}$ is transformed to world coordinates using the rotation matrix $R$, resulting in the following expression:
\begin{equation}
    \covthreed{i}=R\covthreedprime{i}R^\top.
\end{equation}

\subsection{Merging 3D SSGs}
\label{sec:methodology:merging}
\begin{figure*}[b!]
    \centering
    \includegraphics[width=0.8\textwidth]{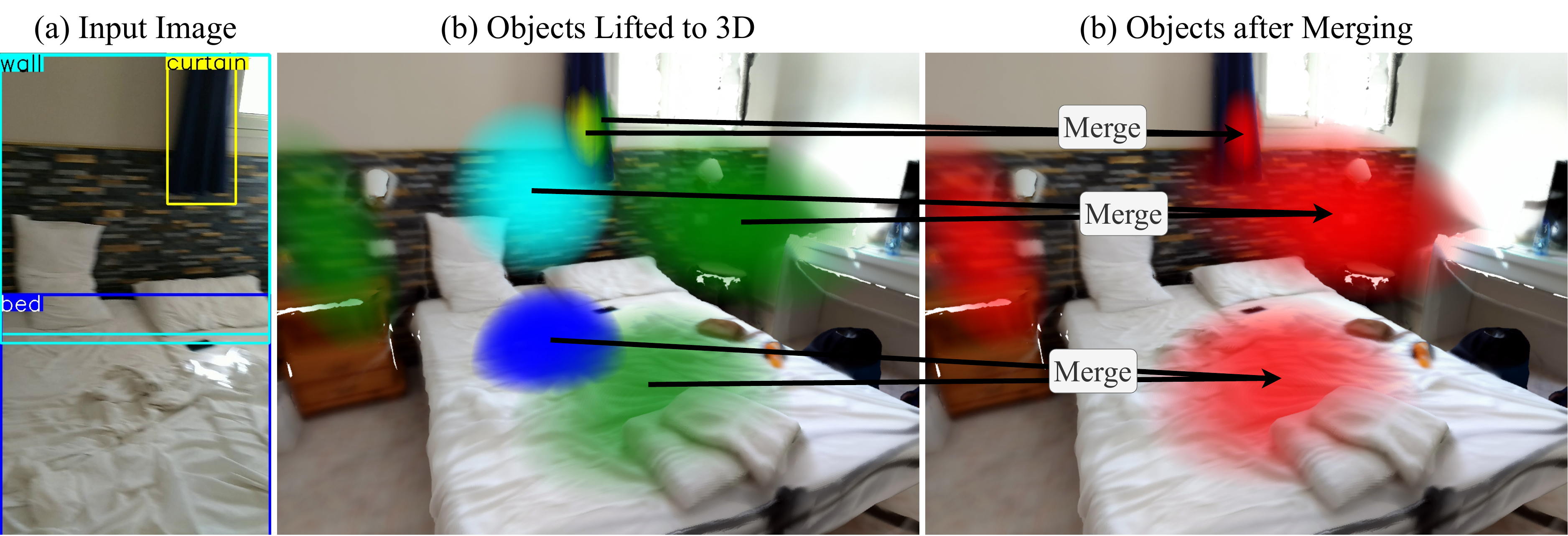}
    \caption{Illustration of the object merging process using the proposed algorithm described in Section~\ref{sec:methodology:merging}. (a) Starting with an input image, a set of objects with predicted categories and bounding boxes is generated. (b) These objects are then lifted into 3D spaces to form the object nodes in $\mathcal{V}_\text{local}$ with colors matching their bounding boxes (\textcolor[RGB]{215, 183, 101}{yellow}, \textcolor{cyan}{cyan}, and \textcolor{blue}{blue}), while pre-existing objects in $\mathcal{V}_\text{global}$ are shown in \textcolor{ForestGreen}{green}. (c) The merging algorithm combines the objects linked by the arrows, resulting in the updated global objects, represented in \textcolor{red}{red}.}
    \label{fig:merging}
\end{figure*}

With a local 3D SSG $\mathcal{G}_\text{local} = (\mathcal{V}_\text{local}, \mathcal{E}_\text{local})$ constructed, it is then integrated with the global 3D SSG $\mathcal{G}_\text{global}$ through Gaussian merging and relationship accumulation. Figure~\ref{fig:merging} illustrates the proposed merging process. Specifically, it initializes a queue with nodes from $\mathcal{V}_\text{local}$ and evaluates potential merging opportunities through distance calculations. 
For each node $v_i$ in this queue, FROSS calculates the Hellinger distances~\cite{10382963} between the Gaussian distributions of $v_i$ and all other nodes $v_j$ of the same semantic class, in both the local node queue and $\mathcal{V}_\text{global}$, to identify object correspondence. The algorithm merges nodes with Hellinger distances below the threshold $\delta_\text{d}$ and adds unmerged nodes to $\mathcal{V}_\text{global}$ as new objects. The Hellinger distance $H_D(i, j)$ between Gaussian distributions $\mathcal{N}(\mu_i, \Sigma_i)$ and $\mathcal{N}(\mu_j, \Sigma_j)$ is quantified through the following equation, wherein $B_D(i, j)$ represents the Bhattacharyya distance:
\begin{align}
    H_D(i, j) &= \sqrt{1-\exp(-B_D(i, j))}, \\
    B_D(i, j) &= \frac{1}{8}\Delta \mu_{ij}^T\Sigma^{-1}\Delta \mu_{ij} + \frac{1}{2}\mathrm{ln}(\frac{\mathrm{det}\Sigma}{\sqrt{\mathrm{det}\Sigma_i\mathrm{det}\Sigma_j}}),
\end{align}
where $\Delta\mu_{ij}=\mu_i-\mu_j$ and $\Sigma=\frac{1}{2}(\Sigma_i+\Sigma_j)$.
The fusion of these nodes occurs through a weighted integration of their respective Gaussian distributions~\cite{crouse2011look}. The weighting coefficients for this integration process reflect both spatial coverage and detection confidence across multiple viewpoints. Objects detected from diverse angular perspectives and spatial positions receive proportionally higher weights, thus mitigating viewpoint bias and ensuring robust three-dimensional representation. Given that weights $w_i$ and $w_j$ represent the detection frequencies of their respective nodes, the mean vector $\mu_k$ and covariance matrix $\Sigma_k$ for the resultant merged node are determined via the formulations:
\begin{align}
    \mu_k &= \frac{w_i\mu_i+w_j\mu_j}{w_i+w_j}, \\
    \Sigma_k &= \frac{w_i\Sigma_i+w_j\Sigma_j}{w_i+w_j} + \frac{w_iw_j(\mu_i-\mu_j)(\mu_i-\mu_j)^\top}{(w_i+w_j)^2},
\end{align}
where $\mu_i$, $\mu_j$ are the mean vectors, and $\Sigma_i$, $\Sigma_j$ are the covariance matrices of the respective Gaussian distributions to be merged. All relationship edges originally connected to $v_i$ and $v_j$ are then redirected to the newly merged node $v_k$. The final relationship prediction between two global object nodes is selected through majority voting, choosing the most frequent prediction across viewpoints.

\section{Experimental Results}
\label{sec:experiment}
The evaluation of FROSS encompasses two 3D SSG datasets: 3DSSG~\cite{wald2020learning} and our developed ReplicaSSG datasets. 
Section~\ref{sec:experiment:setup} introduces the datasets, baseline SSG generation methods, and evaluation metrics. Section~\ref{sec:experiment:implementation} presents the implementation details. Section~\ref{sec:experiment:quantitative} provides comprehensive comparisons between FROSS and existing SSG generation methods on the 3DSSG dataset. Moreover, the effectiveness of our approach is further demonstrated through qualitative results in Section~\ref{sec:experiment:qualitative}. 
We further provide runtime analyses on the ReplicaSSG dataset in Section~\ref{sec:experiment:runtime}, along with additional ablation studies in Section~\ref{sec:experiment:ablation}. More quantitative and qualitative analyses are provided in the supplementary material.

\subsection{Evaluation Setup}
\label{sec:experiment:setup}

\subsubsection{Datasets}
\label{sec:experiment:datasets}

\noindent \textbf{3DSSG.}
The 3DSSG dataset~\cite{wald2020learning} extends 3RScan~\cite{Wald2019RIO}, which encompasses 1,482 scans of indoor environments with their corresponding RGB-D image sequences, 3D meshes, and dense instance segmentation. The dataset contains 363,555 images, with an average of 245 images per scan. 3DSSG augments the base dataset with object attributes, hierarchical category labels, and directed edges that describe inter-object semantic relationships such as `standing on,' `attached to,' and `same color.' This dataset has established a primary benchmark for numerous studies~\cite{wald2020learning, wu2021scenegraphfusion, wu2023incremental, zhang2024egosg, zhang2021exploiting, wang2023vl, zhang2021knowledge, liu2022explore, wald2022learning}. Our experiments utilize the publicly available training and testing splits. To address the class imbalance in both object and relationship labels, this study adopts the commonly used category definitions in~\cite{wu2021scenegraphfusion, wald2020learning, wu2023incremental}. Specifically, the original 160 object categories map to 20 categories from NYUv2~\cite{Silberman:ECCV12}, while seven of the 26 predicate categories remain. Following this categorical mapping, 3DSSG contains a total of 21,974 distinct objects and 16,324 inter-object relationships. This extensive collection of relationships offers a substantial foundation for spatial and semantic analyses. Nevertheless, the low image quality and low frame rate of 3DSSG hinders the comprehensive evaluation of FROSS at its full operational capacity. This necessitates our development of ReplicaSSG, which is an enhanced adaptation of the Replica dataset~\cite{replica19arxiv}.

\noindent \textbf{ReplicaSSG.} 
ReplicaSSG extends the Replica dataset~\cite{replica19arxiv} with newly annotated object relationships. The original Replica provides high-quality reconstructions of indoor environments with instance-segmented meshes and photorealistic renderings, and is particularly suitable for SSG evaluation. Although Replica encompasses only 18 scenes, which precludes its use for training purposes, it serves as an effective evaluation benchmark. In the evaluation of FROSS, seven scenes serve as validation environments for hyperparameter optimization, while the remaining 11 scenes function as performance assessment platforms. Rather than preserving Replica's original object category definitions, ReplicaSSG adopts the classification system from Visual Genome~\cite{krishna2017visual}, a 2D SG dataset. This label mapping from Replica to Visual Genome facilitates zero-shot transfer from 2D SG models, as this approach eliminates the requirement for additional training on the Replica dataset. The dataset splits and label mapping are available in the released code. In addition, detailed statistics for the dataset are provided in the supplementary materials (Section~\ref{sec:supp:statistic}).
As a result, ReplicaSSG offers high-quality instance-segmented meshes with comprehensive relationship annotations.

\subsubsection{Baseline Methods}
We evaluated FROSS on the 3DSSG dataset against several baseline methods for SSG generation. The baselines include three image-based methods, IMP~\cite{xu2017scene}, VGfM~\cite{gay2019visual} and Kim's construction framework~\cite{kim20193}, alongside three point cloud-based methods: 3DSSG~\cite{wald2020learning}, SGFN~\cite{wu2021scenegraphfusion} and the approach presented in~\cite{wu2023incremental}. Kim’s 3D object representation and merging mechanism are integrated into FROSS for its baseline implementation, while other baseline implementations follow~\cite{wu2023incremental}\footnote{\scriptsize{\url{https://github.com/ShunChengWu/3DSSG/tree/4b783ec}}}. In the ablation studies, we investigate the impact of using ground truth 2D SGs and camera trajectories on the ReplicaSSG dataset.

\subsubsection{Matching Object Predictions to Ground Truth}
The evaluation methodology follows~\cite{wu2023incremental} for matching predicted object instances with ground truth counterparts. As FROSS generates predictions without explicit point cloud output, we establish evaluation metrics using back-projected 3D points. These points derive from the central region ($H/2 \times W/2$ pixels) of each object's 2D bounding box. The matching process associates each predicted point with its nearest ground truth point within a 0.1-meter radius. A predicted object establishes correspondence with a ground truth object under two criteria, described as follows:
\begin{enumerate}
    \item The majority (more than 50\%) of its back-projected 3D points must each have their nearest ground truth point neighbor within the corresponding ground truth object.
    \item The ratio between the second-largest and largest ground truth object overlap counts must remain below 75\%.
\end{enumerate}
In contrast to the other baseline methods, which may permit multiple predictions to be matched with a single ground truth object due to possible over-segmentation in input instance-segmented point clouds, our proposed FROSS framework enforces a stricter one-to-one correspondence between predicted and ground truth objects. Since FROSS merges identical objects into a single prediction, matching multiple predictions to the same ground truth object is not required. The one-to-one correspondence enforces the integrity of the proposed merging algorithm, ensuring accurate association between predicted and ground truth objects.

\subsubsection{Evaluation Metrics}
The evaluation of SSG generation employs recall metrics for relationships, objects, and predicate, a standard approach in both 2D and 3D SG evaluation~\cite{johnson2015image, zellers2018neural, wald2020learning, wu2021scenegraphfusion, wu2023incremental}. Object recall quantifies the proportion of ground truth instances matched to predictions with correct category labels. Predicate recall computes the proportion of correctly classified predicates between detected objects, regardless of the object classes. Relationship recall measures the proportion of correctly identified triplets (subject, object, predicate) that meet two criteria: valid detection of both subject and object, and correct classification of their connecting predicate. To mitigate the impact of class imbalance, we report mean recall (mRecall) to provide a more equitable evaluation across all classes. Note that a more detailed explanation of the utilized recall metric is offered in the supplementary materials (Section~\ref{sec:supp:detailed_eval}).

\subsection{Implementation Details}
\label{sec:experiment:implementation}
For our 2D SG generation model, we employ EGTR~\cite{im2024egtr} with RT-DETRv2-M~\cite{zhao2024detrs, lv2024rtdetrv2improvedbaselinebagoffreebies} as the object detection backbone. The 2D SGs from the 3DSSG dataset~\cite{wald2020learning} serve as training data specifically for 3DSSG experiments, while the Visual Genome dataset~\cite{krishna2017visual} functions as the training foundation for ReplicaSSG experiments. Our methodology adheres to the training protocol established in EGTR, which initiates with pre-training the object detector for standard object detection tasks. All hyperparameters were determined through grid search evaluation on the validation split, with particular emphasis on relationship recall optimization. Our implementation applies a confidence threshold of $0.7$ for object filtering and retains only the top ten relationships per 2D SG. For Gaussian merging operations, the Hellinger distance threshold $\delta_\text{d}$ is established at $0.85$. The main experimental evaluations utilize ground truth trajectories to guide the SSG generation process. The impact of estimated trajectories are further analyzed via ablation studies in Section~\ref{sec:experiment:ablation:slam}.

\subsection{Quantitative Results}
\label{sec:experiment:quantitative}

\begin{table}[t]
    \centering
    \caption{
        Performance comparison of 3D SSG generation methods on the 3DSSG dataset and the end-to-end latency without environmental mapping reported in their original literature, along with their respective input modalities. The best and second-best results are highlighted in \textcolor{red}{\textbf{red}}, and \textcolor{blue}{\underline{blue}}, respectively. Latencies are reported in milliseconds.
    }
    \small
    \resizebox{0.475\textwidth}{!}{
        \Large
        \begin{tabular}{l|ccc|cc|c|c}
        \toprule
         & \multicolumn{3}{c|}{\textbf{Recall (\%)}} & \multicolumn{2}{c|}{\textbf{mRecall (\%)}} & & \multirow{2}{*}{\shortstack{\textbf{Input}\\\textbf{Modality}}} \\
        \textbf{Method} & \textbf{Rel.} & \textbf{Obj.} & \textbf{Pred.} & \textbf{Obj.}  & \textbf{Pred.} & \textbf{Latency} & \\
        \hline
        IMP~\cite{xu2017scene} & 19.7 & 49.5 & 20.9 & 34.7 & 13.8 & - & RGB-D \\
        VGfM~\cite{gay2019visual} & 19.6 & 50.0 & 20.4 & 34.8 & 11.0 & 250 & RGB \\
        3DSSG~\cite{wald2020learning} & 12.9 & 37.4 & 22.0 & 26.2 & 14.4 & - & Point Cloud \\ 
        SGFN~\cite{wu2021scenegraphfusion} & 22.0 & 51.6 & 27.5 & 37.7 & \color{blue}{\underline{24.0}} & {\color{blue}\underline{161}} & RGB-D \\
        Wu~\cite{wu2023incremental} & \color{blue}{\underline{23.3}} & 53.8  & \color{blue}{\underline{28.4}} & 43.8 & \color{red}{\textbf{26.6}} & 191 & RGB-D\rlap{ \footnotemark} \\
        Kim~\cite{kim20193} & 9.1 & \color{blue}{\underline{59.0}} & 7.1 & \color{blue}{\underline{51.0}} & 8.0 & 310 & RGB-D \\
        \midrule
        FROSS (Ours) & \color{red}{\textbf{27.9}} & \color{red}{\textbf{62.4}}  & \color{red}{\textbf{33.0}} & \color{red}{\textbf{63.8}} & 18.0 & \color{red}{\textbf{7}} & RGB-D \\
        \bottomrule
        \end{tabular}
    }
    \label{table:main_exp}
\end{table}
\footnotetext{Although Wu's method~\cite{wu2023incremental} supports RGB inputs, the reported results are based on predictions using dense reconstruction from RGB-D images.}

Table~\ref{table:main_exp} presents a performance and latency comparison between FROSS and various representative baseline methodologies. The results reveal that FROSS achieves the highest performance among all baseline methods with much lower processing latency. This substantiates the advantages of lifting scene graphs from 2D images over direct point cloud reasoning~\cite{wald2020learning, wu2021scenegraphfusion, wu2023incremental}, as point clouds can sometimes present challenges with occlusion, sparsity, and reconstruction quality. The superior performance compared to the other image-based methods~\cite{gay2019visual, kim20193, xu2017scene} can be attributed to our Gaussian-based merging strategy. In contrast, IMP~\cite{xu2017scene} employs a voting mechanism for object classification, while VGfM~\cite{gay2019visual} uses a basic recurrent neural network, which may restrict their object and relationship classification accuracy. Kim's baseline~\cite{kim20193} achieves similar object recall to FROSS due to the same 2D SG generation pipeline. However, its merging mechanism fails to suppress duplicate detections, which hinders relationship aggregation and leads to significantly lower relationship and predicate recall. These results collectively substantiate the effectiveness of FROSS in combining 2D information with Gaussian merging for 3D SSG generation.

\subsection{Qualitative Results}
\label{sec:experiment:qualitative}
\begin{figure}[t]
    \centering
    \includegraphics[width=0.475\textwidth]{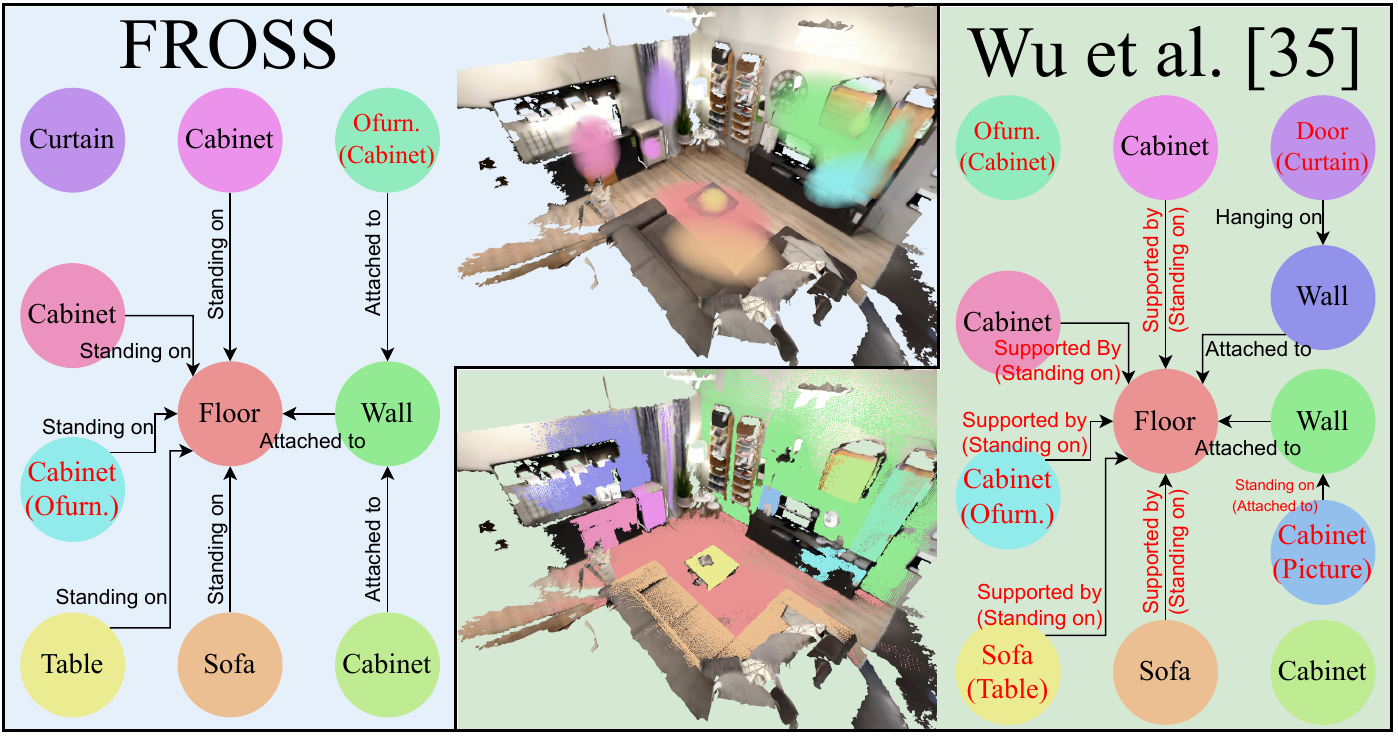}
    \caption{
        Qualitative comparison between FROSS and Wu~\cite{wu2023incremental} on the 3DSSG dataset. Only representative objects are visualized. Errors are marked in \textcolor{red}{red}, with ground truth label shown in parentheses. Node colors correspond to the respective scene graphs.
    }
    \label{fig:qualitative}
\end{figure}
Figure~\ref{fig:qualitative} presents the 3D SSG generation results from FROSS, demonstrating its ability to approximate object positions and shapes within a scene through Gaussian distributions, while simultaneously extracting their relationships. The unified objects highlight the effectiveness of the proposed merging algorithm, which consolidates predictions from multiple views, reducing redundant objects and processing time. This representation method offers a high-level scene understanding for downstream applications. In comparison, Wu's method~\cite{wu2023incremental} frequently misclassifies objects with similar geometry (e.g., ``curtain'' vs. ``door'', ``table'' vs. ``sofa'') and confuses semantically close relationships (e.g., ``standing on'' vs. ``supported by''). This performance gap shows FROSS's advantage in utilizing richer visual information, in contrast to Wu~\cite{wu2023incremental}, which relies solely on cropped object image for visual feature extraction. Additional qualitative results from FROSS are provided in the supplementary materials (Section~\ref{sec:supp:additional_qualitative}).

\subsection{Runtime Analysis}
\label{sec:experiment:runtime}
Table~\ref{table:runtime} examines the latency and frames-per-second (FPS) performance of FROSS, evaluated on an AMD Ryzen 9 7950X processor with an NVIDIA GeForce RTX 4090 GPU. This analysis serves to validate the computational efficiency of FROSS in accordance with its design objective of achieving faster-than-real-time processing capabilities. The evaluation restricts processing to a single CPU core and one GPU to assess system efficiency under limited computational resources. The metrics represent averages across 14,400 frames from four ReplicaSSG test scenes. The RT-DETR object detector and the EGTR relationship extraction module exhibit the highest individual latency. On the other hand, the 3D SSG merging algorithm contributes minimal computational overhead. Compared to the previous methods for online real-time 3D SSG generation methods~\cite{kim20193, wu2021scenegraphfusion, wu2023incremental}, FROSS demonstrates significantly reduced end-to-end latency and increased FPS, which establishes a substantial improvement in terms of speed and efficiency.

\begin{table}[t]
    \centering
    \caption{
        Runtime analysis of the key components of FROSS. \textbf{Obj. Det.} refers to the object detection model RT-DETR~\cite{zhao2024detrs, lv2024rtdetrv2improvedbaselinebagoffreebies}, while \textbf{Rel. Ext.} corresponds to the EGTR relationship extractor~\cite{im2024egtr}. \textbf{Merging 3D SSGs} represents the process described in Section~\ref{sec:methodology:merging}. Only components with significant latency impact are included.
    }
    \footnotesize
    \begin{tabular}{ccc|c}
    \toprule
    \multicolumn{3}{c|}{\textbf{Mean Latency (in millisecond)}} \\
    \textbf{Obj. Det.} & \textbf{Rel. Ext.} & \textbf{Merging 3D SSGs} & \textbf{FPS} \\
    \hline
    2.31 & 4.51 & 0.12 & 144.09 \\
    \bottomrule
    \end{tabular}
    \vspace{-0.5em}
    \label{table:runtime}
\end{table}

\subsection{Ablation Study}
\label{sec:experiment:ablation}

\begin{table}[b]
    \centering
    \caption{
        Analysis of FROSS with predicted and ground truth 2D SGs as input on both the 3DSSG and ReplicaSSG datasets. `w/ GT' denotes the version of FROSS utilizing ground truth 2D SGs.
    }
    \resizebox{0.475\textwidth}{!}{
        \begin{tabular}{l|l|ccc|cc}
        \toprule
         & & \multicolumn{3}{c|}{\textbf{Recall (\%)}} & \multicolumn{2}{c}{\textbf{mRecall (\%)}}\\
        \textbf{Dataset} & \textbf{Method} & \textbf{Rel.} & \textbf{Obj.} & \textbf{Pred.} & \textbf{Obj.} & \textbf{Pred.} \\
        \hline
        \multirow{2}{*}{3DSSG~\cite{wald2020learning}} & FROSS & 27.9 & 62.4 & 33.0 & 63.8 & 18.0 \\
         & FROSS w/ GT & 55.8 & 88.6 & 56.0 & 93.3 & 56.8 \\
        \midrule
        \multirow{2}{*}{ReplicaSSG} & FROSS & 22.3 & 26.1 & 27.8 & 28.8 & 20.4 \\
         & FROSS w/ GT & 67.6 & 89.1 & 67.6 & 92.6 & 82.6 \\
        \bottomrule
        \end{tabular}
    }
    \label{table:gt}
\end{table}

\subsubsection{Using Ground Truth 2D Scene Graphs}
\label{sec:experiment:gt2dsg}
In the ablation analysis presented in Table~\ref{table:gt}, we investigate the influence of 2D SG quality on FROSS performance, given that FROSS relies heavily on accurate 2D SG predictions to generate 3D SSGs. The results indicate that the accuracy of the 2D SG has a significant impact on the final 3D SSG quality. In both the 3DSSG and ReplicaSSG datasets, FROSS achieves substantial gains in recall when ground truth 2D SG is used instead of predicted 2D SG. This performance difference is particularly pronounced in the ReplicaSSG dataset, where the baseline FROSS method achieves a relatively low recall score compared to the ground truth-assisted version. The discrepancy stems from differences between ReplicaSSG and Visual Genome~\cite{krishna2017visual}. Since the 2D SG model was originally trained on Visual Genome, the model may not generalize effectively to datasets with different characteristics. This leads to lower recall in both object and relationship detection within the ReplicaSSG dataset. This finding suggests the untapped potential for FROSS via domain-adaptive 2D SG models, with current results merely representing a lower bound on the framework's capabilities.

\subsubsection{Using Estimated Camera Trajectory}
\label{sec:experiment:ablation:slam}
We further validate FROSS's effectiveness using estimated camera trajectories instead of ground truth poses to assess real-world applicability. This evaluation is crucial as most practical deployments must rely on estimated camera poses rather than perfect ground truth information. Table~\ref{table:slam_traj} presents the impact of using estimated trajectories generated by ORB-SLAM3~\cite{campos2021orb} on the ReplicaSSG dataset. The evaluation results demonstrate a slight improvement in relationship recall, while object and predicate recall exhibit minor decreases. This indicates that FROSS remains robust to inaccuracies in camera trajectory estimation. Note that a more detailed analysis of ORB-SLAM3's trajectory estimation errors on the ReplicaSSG dataset is provided for reference in the supplementary materials~(Section~\ref{sec:supp:orbslam3}).

\begin{table}[t]
    \centering
    \caption{
        Performance comparison of FROSS with the predicted and the ground truth camera trajectories as input on the ReplicaSSG dataset. Please note that `w/ SLAM' denotes the version of FROSS utilizing trajectories estimated by ORB-SLAM3~\cite{campos2021orb}.
    }
    \footnotesize
    \resizebox{0.475\textwidth}{!}{
        \begin{tabular}{l|ccc|cc}
        \toprule
         & \multicolumn{3}{c|}{\textbf{Recall (\%)}} & \multicolumn{2}{c}{\textbf{mRecall (\%)}} \\
        \textbf{Method} & \textbf{Rel.} & \textbf{Obj.} & \textbf{Pred.} & \textbf{Obj.} & \textbf{Pred.}\\
        \hline
        FROSS & 22.3 & 26.1 & 27.8 & 28.8 & 20.4 \\
        FROSS w/ SLAM & 22.7 & 25.8 & 27.2 & 27.7 & 20.1 \\
        \bottomrule
        \end{tabular}
    }
    \label{table:slam_traj}
\end{table}

\subsubsection{Effect of Hellinger Distance Threshold}
\label{sec:experiment:ablation:hellinger}
Table~\ref{table:hellinger} presents the effect of varying the Hellinger distance threshold $\delta_\text{d}$. A lower threshold (e.g., 0.65) enforces stricter merging criteria, which preserves object distinctions and leads to higher object recall. In contrast, a higher threshold (e.g., 0.85) enables more extensive relationship aggregation between objects, which improves relationship performance.

\begin{table}[t]
    \centering
    \caption{Recall on the validation split of the 3DSSG dataset with different Hellinger distance thresholds.}
    \resizebox{0.475\textwidth}{!}{
    \begin{tabular}{c|c|ccccccc}
        \toprule
        \multicolumn{2}{c|}{\textbf{Threshold $\delta_\text{d}$}} & 0.6 & 0.65 & 0.7 & 0.75 & 0.8 & 0.85 & 0.9 \\
        \hline
        \multirow{3}{*}{Recall (\%)} & \textbf{Rel.} & 15.1 & 16.5 & 20.0 & 22.7 & 25.9 & \color{red}{\textbf{30.0}} & \color{blue}{\underline{29.3}} \\
        & \textbf{Obj.} & \color{blue}{\underline{69.5}} & \color{red}{\textbf{69.8}} & 69.1 & 68.9 & 65.7 & 61.9 & 55.4 \\
        & \textbf{Pred.} & 16.2 & 17.9 & 21.8 & 24.9 & 28.8 & \color{blue}{\underline{33.6}} & \color{red}{\textbf{34.3}} \\
        \bottomrule
    \end{tabular}
    }
    \vspace{-0.5em}
    \label{table:hellinger}
\end{table}

\section{Conclusion}
\label{sec:conclusion}
This study presented FROSS, an innovative framework for real-time generation of 3D SSG in online environments. By representing objects as 3D Gaussian distributions, FROSS achieved computational efficiency that exceeded real-time requirements. The framework integrated a real-time object detection and relationship extraction pipeline to generate 2D SGs, which underwent 3D back-projection and incremental integration into a global SSG through our proposed merging algorithm. Experimental evaluations on existing benchmark and our newly proposed dataset, ReplicaSSG, demonstrated that FROSS surpassed existing methods in both performance metrics and computational speed by significant margins. These advancements established possibilities for real-time SSG applications in open-world environments, particularly relevant to robotics and AR domains.

\newpage

\section*{Acknowledgments}
This work was partly supported by the Japan Society for the Promotion of Science KAKENHI Grant Numbers JP24H00733. The authors gratefully acknowledge the support from the National Science and Technology Council (NSTC) in Taiwan under grant numbers NSTC 114-2221-E-002-069-MY3, NSTC 113-2221-E-002-212-MY3, NSTC 114-2218-E-A49-026, and NSTC 114-2640-E-002-006. The authors would also like to express their appreciation for the donation of the GPUs from NVIDIA Corporation and NVIDIA AI Technology Center (NVAITC) used in this work. Furthermore, the authors extend their gratitude to the National Center for High-Performance Computing for providing the necessary computational and storage resources.

{
    \small
    \bibliographystyle{ieeenat_fullname}
    \bibliography{main}
}

\clearpage
\setcounter{page}{1}
\maketitlesupplementary

\begin{table*}[b]
    \centering
    \caption{
        Per-class performance comparison of 3D SSG generation methods on 3DSSG for object recall (\%). The best and second-best results are highlighted in \textcolor{red}{\textbf{red}}, and \textcolor{blue}{\underline{blue}}, respectively.
    }
    \resizebox{\textwidth}{!}{
        \LARGE
        \begin{tabular}{l|cccccccccccccccccccc|c}
            \toprule
            \textbf{Method} & \textbf{bath.} & \textbf{bed} & \textbf{bkshf.} & \textbf{cab.} & \textbf{chair} & \textbf{cntr.} & \textbf{curt.} & \textbf{desk} & \textbf{door} & \textbf{floor} & \textbf{ofurn.} & \textbf{pic.} & \textbf{refri.} & \textbf{show.} & \textbf{sink} & \textbf{sofa} & \textbf{table.} & \textbf{toil.} & \textbf{wall} & \textbf{wind.} & \textbf{mean} \\
            \hline
            IMP~\cite{xu2017scene} & 0.0 & 66.7 & 0.0 & 38.1 & 45.3 & 0.0 & 47.7 & 0.0 & 8.1 & 95.1 & 19.9 & 2.3 & 0.0 & 0.0 & 20.0 & 47.4 & 48.5 & \textcolor{blue}{\underline{66.7}} & \textcolor{blue}{\underline{77.0}} & \textcolor{blue}{\underline{17.9}} & 30.0 \\
            VGfM~\cite{gay2019visual} & 0.0 & 66.7 & 0.0 & 34.6 & 49.4 & 0.0 & 48.6 & 4.2 & 19.8 & \textcolor{blue}{\underline{95.7}} & 14.1 & 1.1 & 0.0 & 0.0 & 23.3 & \textcolor{blue}{\underline{57.9}} & 56.9 & 63.0 & \color{red}{\textbf{78.0}} & \textcolor{blue}{\underline{17.9}} & 31.6 \\
            3DSSG~\cite{wald2020learning} & 25.0 & 66.7 & 0.0 & 20.0 & 51.0 & \textcolor{blue}{\underline{25.8}} & \textcolor{blue}{\underline{50.5}} & 0.0 & \textcolor{blue}{\underline{47.7}} & 91.4 & 14.7 & 3.4 & \textcolor{blue}{\underline{22.2}} & \textcolor{blue}{\underline{14.3}} & 25.0 & 47.4 & 42.5 & 25.9 & 51.9 & 13.1 & 31.9 \\
            SGFN~\cite{wu2021scenegraphfusion} & \textcolor{blue}{\underline{75.0}} & 33.3 & 0.0 & \textcolor{blue}{\underline{50.8}} & 63.6 & 19.4 & 40.5 & 8.3 & 38.7 & \color{red}{\textbf{96.9}} & 23.0 & \textcolor{blue}{\underline{11.4}} & 11.1 & 0.0 & \textcolor{blue}{\underline{38.3}} & 55.3 & \textcolor{blue}{\underline{62.3}} & 51.9 & 73.0 & 13.1 & 38.3 \\
            Wu~\cite{wu2023incremental} & \textcolor{blue}{\underline{75.0}} & \color{red}{\textbf{100.0}} & 0.0 & 50.4 & \color{red}{\textbf{65.6}} & 19.4 & 45.9 & \textcolor{blue}{\underline{12.5}} & 34.2 & \color{red}{\textbf{96.9}} & \textcolor{blue}{\underline{25.1}} & 5.7 & 0.0 & \textcolor{blue}{\underline{14.3}} & \textcolor{blue}{\underline{38.3}} & \textcolor{blue}{\underline{57.9}} & 59.9 & \textcolor{blue}{\underline{66.7}} & 76.1 & 15.5 & \textcolor{blue}{\underline{43.0}} \\
            \midrule
            FROSS (Ours) & \color{red}{\textbf{100.0}} & \textcolor{blue}{\underline{83.3}}  & \color{red}{\textbf{28.6}} & \color{red}{\textbf{56.1}} & \textcolor{blue}{\underline{64.8}} & \color{red}{\textbf{67.7}} & \color{red}{\textbf{73.0}} & \color{red}{\textbf{29.2}} & \color{red}{\textbf{73.3}} & 91.5 & \color{red}{\textbf{40.3}} & \color{red}{\textbf{41.9}} & \color{red}{\textbf{50.0}} & \color{red}{\textbf{42.9}} & \color{red}{\textbf{73.3}} & \color{red}{\textbf{73.7}} & \color{red}{\textbf{68.2}} & \color{red}{\textbf{100.0}} & 60.9 & \color{red}{\textbf{57.5}} & \color{red}{\textbf{63.8}} \\
            \bottomrule
        \end{tabular}
    }
    \vspace{-0.5em}
    \label{table:supp:obj_per_class}
\end{table*}

\begin{table*}[b]
    \centering
    \caption{
        Per-class performance comparison of 3D SSG generation methods on 3DSSG for predicate recall (\%). The best and second-best results are highlighted in \textcolor{red}{\textbf{red}}, and \textcolor{blue}{\underline{blue}}, respectively.
    }
    \resizebox{\textwidth}{!}{
        \scriptsize
        \begin{tabular}{l|ccccccc|c}
            \toprule
            \textbf{Method} & \textbf{attached to} & \textbf{build in} & \textbf{connected to} & \textbf{hanging on} & \textbf{part of} & \textbf{standing on} & \textbf{supported by} & \textbf{mean} \\
            \hline
            IMP~\cite{xu2017scene} & 48.4 & 7.7 & 21.7 & \textcolor{blue}{\underline{11.9}} & 0.0 & 1.4 & 5.5 & 13.8 \\
            VGfM~\cite{gay2019visual} & 49.1 & 2.6 & 10.9 & 5.2 & 0.0 & 0.5 & 8.8 & 11.0 \\
            3DSSG~\cite{wald2020learning} & 46.6 & 15.4 & 10.9 & \textcolor{blue}{\underline{11.9}} & 0.0 & \textcolor{blue}{\underline{1.8}} & 14.3 & 14.4 \\
            SGFN~\cite{wu2021scenegraphfusion} & \color{red}{\textbf{58.4}} & \textcolor{blue}{\underline{33.3}} & \textcolor{blue}{\underline{32.6}} & \color{red}{\textbf{26.1}} & 0.0 & 1.0 & \color{red}{\textbf{16.5}} & \textcolor{blue}{\underline{24.0}} \\
            Wu~\cite{wu2023incremental} & \textcolor{blue}{\underline{58.0}} & \textcolor{blue}{\underline{33.3}} & \color{red}{\textbf{39.1}} & \color{red}{\textbf{26.1}} & \color{red}{\textbf{12.5}} & 1.5 & \textcolor{blue}{\underline{15.4}} & \color{red}{\textbf{26.6}} \\
            \midrule
            FROSS (Ours) & 29.4 & \color{red}{\textbf{43.6}} & 0.0 & 1.4 & 0.0 & \color{red}{\textbf{47.2}} & 4.2 & 18.0 \\
            \bottomrule
        \end{tabular}
    }
    \label{table:supp:pred_per_class}
\end{table*}

\section{Detailed Evaluation Metric}
\label{sec:supp:detailed_eval}
The evaluation procedure in this paper follows closely with Wu~\cite{wu2023incremental} to ensure a fair comparison. The only difference is the exclusion of the `none' relationship category, as FROSS does not predict it. Wu~\cite{wu2023incremental} also provided results evaluated under this protocol in their publicly released code.

It is important to clarify that the \textit{recall@1} metric stated in~\cite{wu2023incremental} differs from the conventional \textit{recall@N} metric used in 2D SG evaluation~\cite{im2024egtr, xu2017scene, zellers2018neural}. In standard \textit{recall@N} evaluation, only the top-N relationship triplets with the highest confidence scores are considered. In contrast, the \textit{recall@1} metric employed in our work and~\cite{wu2023incremental} focuses solely on the predicted class labels within a detected triplet. Specifically, for a detected triplet in which both the subject and object match ground truth objects, only the predicted class labels for the subject, object, and predicate with the highest confidence scores are considered. Notably, as our approach does not impose a restriction on the number of detected relationship triplets, the \textit{recall@1} metric in~\cite{wu2023incremental} is conceptually more aligned with \textit{recall@$\infty$} with graph constraints~\cite{zellers2018neural}. To mitigate potential confusion, we refer to this metric simply as \textit{recall} throughout our work.

Additionally, the predicate recall metric used in this study does not fully correspond to the conventional predicate classification (\textbf{PredCls})~\cite{xu2017scene} setting, as no ground truth objects are provided.

\section{Additional Experimental Results}
\label{sec:supp:additional_exp}

\subsection{Object and Predicate Performance per Class}
The per-class performance comparison of FROSS and other baselines is presented in Tables~\ref{table:supp:obj_per_class} and~\ref{table:supp:pred_per_class}. In addition, FROSS's per-class object and predicate performance on the proposed ReplicaSSG dataset is presented in Table~\ref{table:supp:obj_per_class_replicassg}.

FROSS excels in detecting object classes that rely heavily on visual information, particularly those with similar geometric structures, such as \textit{bookshelf}, \textit{counter}, \textit{desk}, \textit{picture}, \textit{refrigerator}, \textit{shower curtain}, and \textit{window}. These objects are often box-like or flat. FROSS’s ability to capture complex visual features leads to significantly higher performance in both object recall and mean recall.

FROSS’s predicate performance is significantly affected by class imbalance, excelling in relationship classes such as \textit{attached to}, \textit{build in}, and \textit{standing on}, while performing poorly on others. Despite retaining only the top seven most frequent relationships, the 3DSSG dataset still exhibits an extreme imbalance, with the top two classes occurring at substantially higher frequencies than the others~\cite{wu2023incremental}. While addressing this issue could potentially enhance FROSS’s performance, we leave it as future work, as class imbalance is not the primary focus of this research.

\subsection{Additional Qualitative Results}
\label{sec:supp:additional_qualitative}
\begin{figure*}[t]
    \centering
    \includegraphics[width=0.925\textwidth]{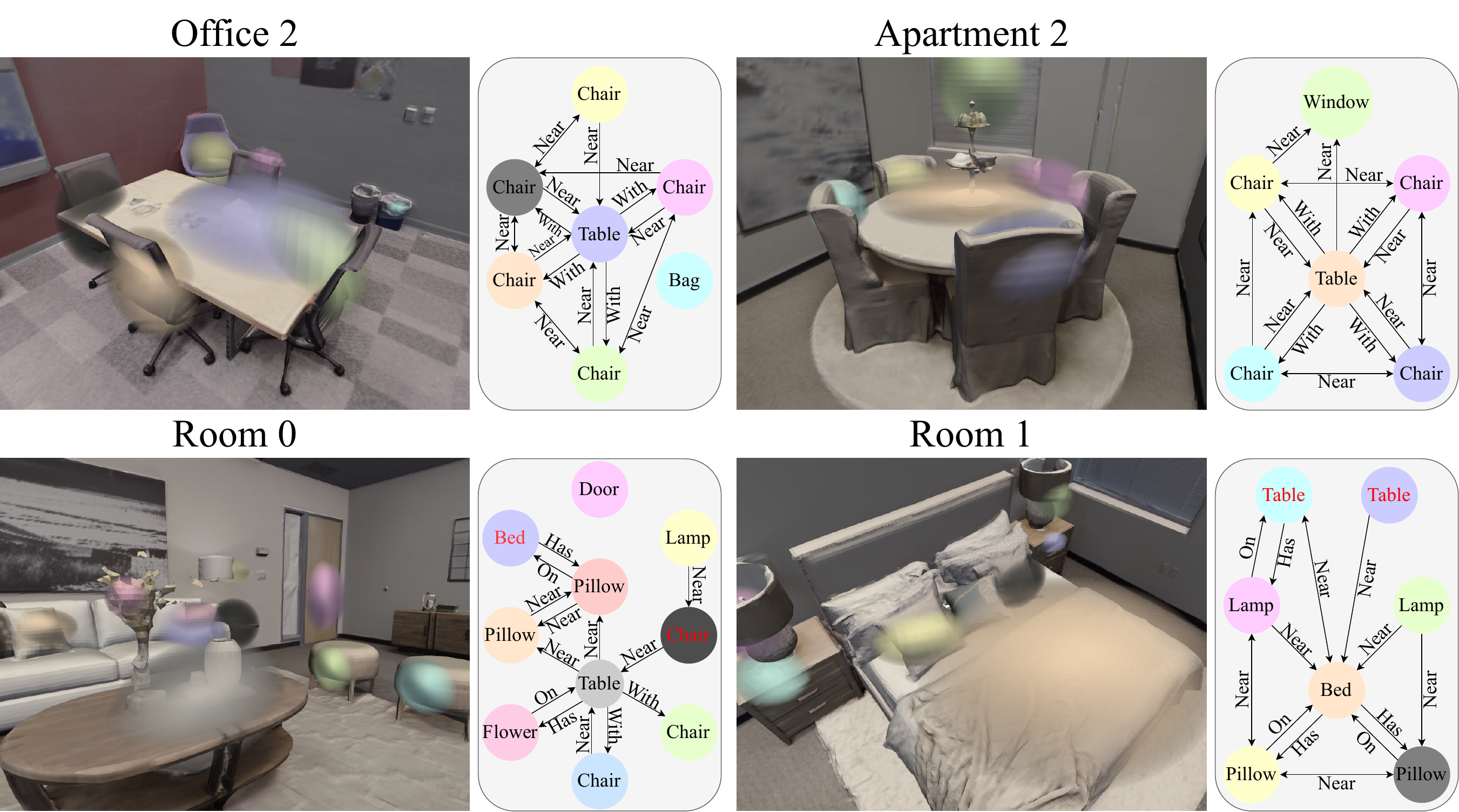}
    \caption{
        Qualitative results of FROSS on four scenes in the ReplicaSSG dataset. Please note that only representative objects are visualized, with misclassified objects marked in \textcolor{red}{red} on the right graph. The node colors in the left visualization correspond to the graphs on the right.
    }
    \label{fig:additional_qualitative}
\end{figure*}

Figure~\ref{fig:additional_qualitative} presents 3D SSG generation results from FROSS on the ReplicaSSG dataset. FROSS captures both spatial relationships (e.g., ``near'' and ``on'') and semantic relationships (e.g., ``has'' and ``with''). Misclassified objects are likely caused by occlusions from certain viewpoints or unusual viewing angles. These results further demonstrate FROSS’s robustness in diverse scene conditions.

\subsection{2D Scene Graph Generation Performance}
In this section, we present the evaluation of two models: the original EGTR~\cite{im2024egtr} 2D SG generation model and our modified version employed in FROSS, \textit{RT-DETR+EGTR}. The latter replaces the object detection backbone in the original EGTR with RT-DETR~\cite{zhao2024detrs} object detector. These models are assessed on three datasets: Visual Genome~\cite{krishna2017visual}, 3DSSG~\cite{wald2020learning}, and the proposed ReplicaSSG, as detailed in Table~\ref{table:2DSG}. For these evaluations, the models tested on ReplicaSSG received training on the Visual Genome dataset, whereas the models tested on the other two datasets used their respective training splits. Moreover, both models were optimized and accelerated using TensorRT\footnote{\url{https://github.com/NVIDIA/TensorRT}}. The evaluation results demonstrate that \textit{RT-DETR+EGTR} achieves superior performance in object detection (AP@50) and decreases processing latency by more than half. On the other hand, the original EGTR model demonstrates better performance in relationship prediction tasks. The above observations reveal that the integration of RT-DETR as the object detection backbone results in substantial processing efficiency improvements, with only a slight impact on relationship prediction performance for the ReplicaSSG dataset. This trade-off highlights the potential of RT-DETR in enhancing EGTR's practicality for applications that require faster inference speed. Moreover, the per-class object and predicate performance of \textit{RT-DETR+EGTR} are shown in Tables~\ref{table:supp:obj_per_class_2dsg} and~\ref{table:supp:rel_per_class_2dsg}.

\subsection{ORB-SLAM3 Performance on ReplicaSSG}
\label{sec:supp:orbslam3}
Table~\ref{table:supp:orbslam_ate} presents the root mean square absolute trajectory error (RMS ATE) for ORB-SLAM3~\cite{campos2021orb} on the proposed ReplicaSSG dataset. The evaluation is conducted using ORB-SLAM3 with its default parameters and RGB-D input. The results are consistent with those reported in the original literature, confirming that ORB-SLAM3 can reliably track trajectories within the ReplicaSSG dataset.

\section{Statistics of the ReplicaSSG Dataset}
\label{sec:supp:statistic}
\begin{figure}[t]
    \centering
    \includegraphics[width=0.475\textwidth]{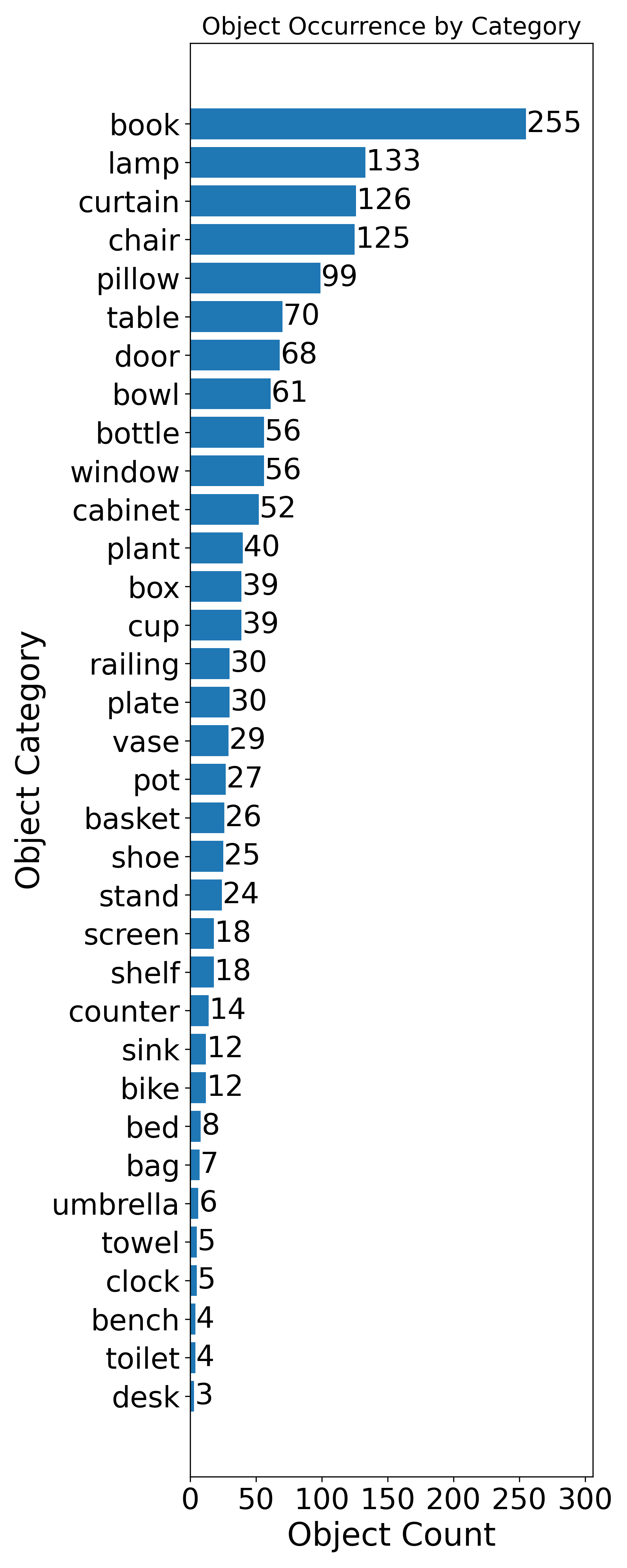}
    \caption{
        The occurrence frequency of each object category in the ReplicaSSG dataset.
    }
    \label{fig:obj_occ}
\end{figure}

\begin{figure}[t]
    \centering
    \includegraphics[width=0.475\textwidth]{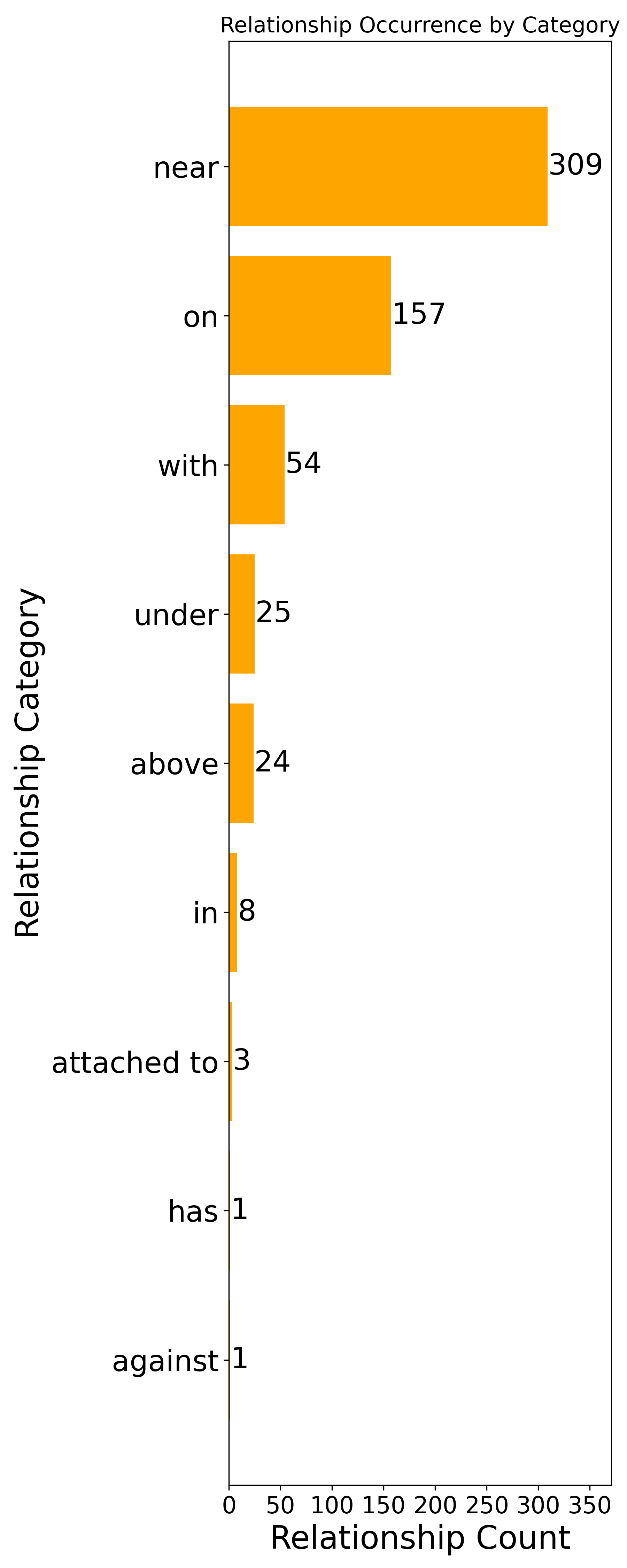}
    \caption{
        The occurrence frequency of each relationship category in the ReplicaSSG dataset.
    }
    \label{fig:rel_occ}
\end{figure}

\begin{figure}[t]
    \centering
    \includegraphics[width=0.475\textwidth]{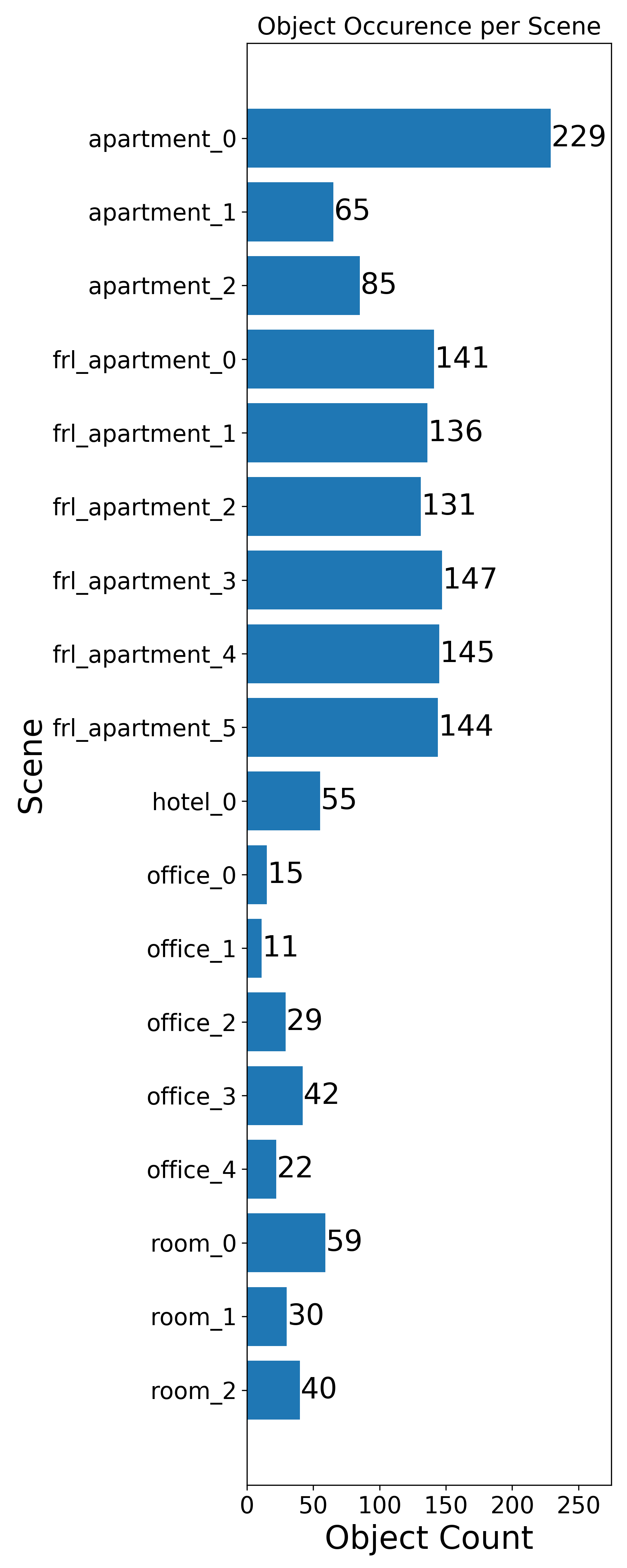}
    \caption{
        The number of objects present in each scene within the ReplicaSSG dataset.
    }
    \label{fig:obj_occ_per_scene}
\end{figure}

\begin{figure}[t]
    \centering
    \includegraphics[width=0.475\textwidth]{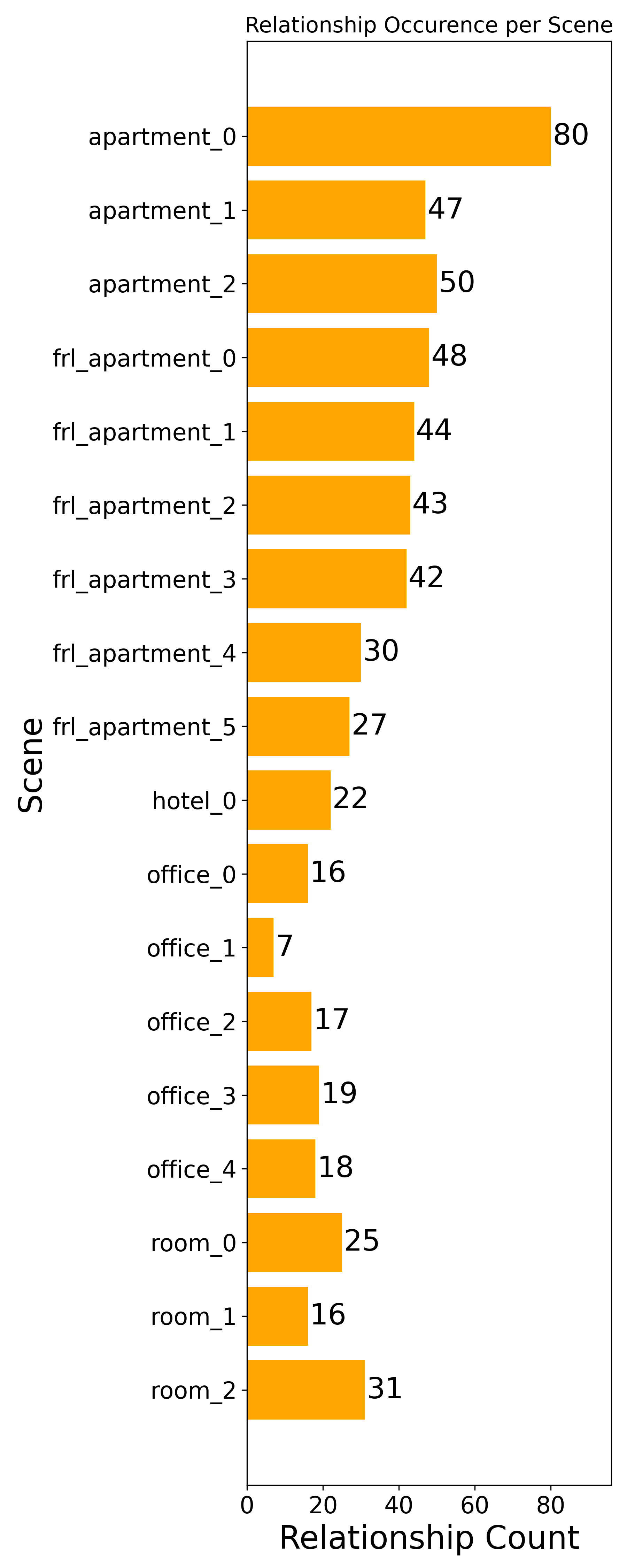}
    \caption{
        The number of relationships present in each scene within the ReplicaSSG dataset.
    }
    \label{fig:rel_occ_per_scene}
\end{figure}

The statistics of the proposed ReplicaSSG Dataset are presented in Figures~\ref{fig:obj_occ}-\ref{fig:rel_occ_per_scene}. More specifically, Figure~\ref{fig:obj_occ} and~\ref{fig:rel_occ} illustrate the occurrence frequency of objects and relationships across all categories in the dataset. In addition, Figures~\ref{fig:obj_occ_per_scene} and~\ref{fig:rel_occ_per_scene} offer scene-specific statistics that detail the number of objects and relationships in each scene.

\begin{table*}[t]
    \centering
    \caption{
        Per-class performance comparison of FROSS on the ReplicaSSG dataset for object and predicate recall (\%).
    }
    \vspace{-0.5em}
    \resizebox{\textwidth}{!}{
        \large
        \begin{tabular}{ccccccccccccccccc|c}
            \toprule
            \multicolumn{18}{c}{\textbf{Object Recall per Class}} \\
            \midrule
            \textbf{bag} & \textbf{bskt.} & \textbf{bed} & \textbf{bench} & \textbf{bike} & \textbf{book} & \textbf{botl.} & \textbf{bowl} & \textbf{box} & \textbf{cab.} & \textbf{chair} & \textbf{clock} & \textbf{cntr.} & \textbf{cup} & \textbf{curt.} & \textbf{desk} & \textbf{door} & \textbf{mean} \\
            25.0 & 50.0 & 0.0 & 0.0 & 0.0 & 1.5 & 9.1 & 37.5 & 4.0 & 14.3 & 68.1 & 66.7 & 40.0 & 33.3 & 9.1 & 0.0 & 80.0 & \multirow{3}{*}{28.8} \\
            \textbf{lamp} & \textbf{pil.} & \textbf{plant} & \textbf{plate} & \textbf{pot} & \textbf{rail.} & \textbf{scrn.} & \textbf{shlf.} & \textbf{shoe} & \textbf{sink} & \textbf{stand} & \textbf{table} & \textbf{toil.} & \textbf{towel} & \textbf{umb.} & \textbf{vase} & \textbf{wind.} \\
            16.7 & 41.5 & 47.4 & 31.2 & 7.7 & 0.0 & 0.0 & 11.1 & 8.3 & 100.0 & 0.0 & 72.2 & 100.0 & 0.0 & 66.7 & 38.9 & 0.0 & \\
            \midrule
            \multicolumn{18}{c}{\textbf{Predicate Recall per Class}} \\
            \midrule
            \multicolumn{2}{c}{\textbf{above}} & \multicolumn{2}{c}{\textbf{against}} & \multicolumn{2}{c}{\textbf{attached to}} & \multicolumn{2}{c}{\textbf{in}} & \multicolumn{2}{c}{\textbf{near}} & \multicolumn{2}{c}{\textbf{on}} & \multicolumn{2}{c}{\textbf{under}} & \multicolumn{2}{c}{\textbf{with}} & & {\textbf{mean}} \\
            \multicolumn{2}{c}{22.2} & \multicolumn{2}{c}{0.0} & \multicolumn{2}{c}{0.0} & \multicolumn{2}{c}{33.3} & \multicolumn{2}{c}{28.8} & \multicolumn{2}{c}{19.1} & \multicolumn{2}{c}{10.0} & \multicolumn{2}{c}{50.0} & & 20.4 \\
            \bottomrule
        \end{tabular}
    }
    \label{table:supp:obj_per_class_replicassg}
\end{table*}

\begin{table*}[b]
    \centering
    \caption{
        Evaluation results of two 2D SG generation models across three datasets. `RT-DETR+EGTR' represents the EGTR model with RT-DETR as its object detector backbone. Latencies are reported in milliseconds. Recall@K (denoted as R@K) provides the class-agnostic average recall, while mean Recall@K (denoted as mR@K) represents the average recall across all relationship categories. All relationship metrics are evaluated with graph constraints as described in~\cite{zellers2018neural}.
    }
    \vspace{-0.5em}
    \resizebox{1.0\textwidth}{!}{
        \begin{tabular}{l|c|c|c|ccc|ccc}
            \toprule
             & & & & \multicolumn{6}{c}{\textbf{Relationship}} \\
            \textbf{Dataset} & \textbf{Method} & \textbf{Latency} & \textbf{AP@50} & \textbf{R@20} & \textbf{R@50} & \textbf{R@100} & \textbf{mR@20} & \textbf{mR@50} & \textbf{mR@100} \\
            \hline
            \multirow{2}{*}{Visual Genome~\cite{krishna2017visual}} & EGTR & 14.6 & 30.8 & 23.5 & 30.2 & 34.3 & 5.5 & 7.9 & 10.1 \\
             & RT-DETR+EGTR & 6.82 & 32.2 & 15.7 & 22.0 & 26.6 & 3.3 & 4.9 & 6.2 \\
            \midrule
            \multirow{2}{*}{ReplicaSSG} & EGTR & 14.6 & 21.2 & 13.2 & 18.1 & 22.0 & 6.9 & 9.6 & 11.9 \\
             & RT-DETR+EGTR & 6.82 & 23.8 & 12.4 & 17.1 & 21.0 & 6.5 & 9.1 & 11.2 \\
            \midrule
            3DSSG~\cite{wald2020learning} & RT-DETR+EGTR & 6.82 & 41.0 & 43.4 & 48.2 & 52.0 & 23.3 & 27.3 & 30.7 \\
            \bottomrule
        \end{tabular}
    }
    \label{table:2DSG}
\end{table*}

\begin{table*}[b]
    \centering
    \caption{
        Per-class object detection performance in 2D SG generation with RT-DETR (AP@50).
    }
    \vspace{-0.5em}
    \resizebox{\textwidth}{!}{
        \Huge
        \begin{tabular}{l|cccccccccccccccccccc|c}
            \toprule
            \textbf{Dataset} & \multicolumn{21}{c}{\textbf{Object Detection AP@50 per Class}} \\
            \hline
            \multirow{2}{*}{3DSSG~\cite{wald2020learning}} & \textbf{bath.} & \textbf{bed} & \textbf{bkshf.} & \textbf{cab.} & \textbf{chair} & \textbf{cntr.} & \textbf{curt.} & \textbf{desk} & \textbf{door} & \textbf{floor} & \textbf{ofurn.} & \textbf{pic.} & \textbf{refri.} & \textbf{show.} & \textbf{sink} & \textbf{sofa} & \textbf{table.} & \textbf{toil.} & \textbf{wall} & \textbf{wind.} & \textbf{mAP@50} \\
            & 100.0 & 83.3  & 28.6 & 56.1 & 64.8 & 67.7 & 73.0 & 29.2 & 73.3 & 91.5 & 40.3 & 41.9 & 50.0 & 42.9 & 73.3 & 73.7 & 68.2 & 100.0 & 60.9 & 57.5 & 63.8 \\
            \midrule
            \multirow{4}{*}{ReplicaSSG} & \textbf{bag} & \textbf{bskt.} & \textbf{bed} & \textbf{bench} & \textbf{bike} & \textbf{book} & \textbf{botl.} & \textbf{bowl} & \textbf{box} & \textbf{cab.} & \textbf{chair} & \textbf{clock} & \textbf{cntr.} & \textbf{cup} & \textbf{curt.} & \textbf{desk} & \textbf{door} & \textbf{lamp} & \textbf{pil.} & \textbf{plant} & \textbf{mAP@50} \\
            & 1.0 & 13.2 & 21.6 & 6.8 & 52.9 & 5.1 & 4.9 & 15.6 & 1.4 & 28.9 & 50.2 & 47.7 & 47.8 & 6.3 & 5.2 & 5.5 & 39.4 & 21.5 & 62.4 & 45.4 & \multirow{3}{*}{23.8} \\
            & \textbf{plate} & \textbf{pot} & \textbf{rail.} & \textbf{scrn.} & \textbf{shlf.} & \textbf{shoe} & \textbf{sink} & \textbf{stand} & \textbf{table} & \textbf{toil.} & \textbf{towel} & \textbf{umb.} & \textbf{vase} & \textbf{wind.}
             & & & & & & \\
            & 14.0 & 3.4 & 18.9 & 34.1 & 24.5 & 4.1 & 43.5 & 0.7 & 44.3 & 61.4 & 2.3 & 7.5 & 30.3 & 37.0 & & & & & & \\
            \bottomrule
        \end{tabular}
    }
    \label{table:supp:obj_per_class_2dsg}
\end{table*}

\begin{table*}[b]
    \centering
    \caption{
        Per-class relationship extraction performance in 2D SG generation with RT-DETR+EGTR (Recall@K).
    }
    \vspace{-0.5em}
    \resizebox{\textwidth}{!}{
        \huge
        \begin{tabular}{l|c|ccccccccc|c}
            \toprule
            \textbf{Dataset} & \textbf{Recall@K} & \multicolumn{9}{c}{\textbf{Relationship Recall@K per Class}} \\
            \hline
            \multirow{4}{*}{3DSSG~\cite{wald2020learning}} & & \textbf{attached to} & \textbf{build in} & \textbf{connected to} & \textbf{hanging on} & \textbf{part of} & \textbf{standing on} & \textbf{supported by} & & & \textbf{mean} \\
            & Recall@20 & 55.7 & 32.1 & 0.3 & 7.0 & 5.4 & 54.1 & 8.7 & & & 23.3 \\
            & Recall@50 & 61.6 & 37.6 & 1.4 & 9.1 & 11.3 & 59.5 & 10.2 & & & 27.3\\
            & Recall@100 & 65.2 & 42.3 & 2.1 & 11.3 & 18.4 & 64.2 & 11.0 & & & 30.7\\
            \midrule
            \multirow{4}{*}{ReplicaSSG} & & \textbf{above} & \textbf{against} & \textbf{attached to} & \textbf{has} & \textbf{in} & \textbf{near} & \textbf{on} & \textbf{under} & \textbf{with} & \textbf{mean} \\
            & Recall@20 & 2.1 & 0.0 & 0.0 & 0.0 & 13.2 & 11.2 & 13.1 & 0.0 & 19.1 & 6.5 \\
            & Recall@50 & 4.4 & 0.0 & 0.0 & 0.0 & 15.3 & 15.8 & 17.0 & 0.0 & 29.2 & 9.1 \\
            & Recall@100 & 6.8 & 0.0 & 0.0 & 0.0 & 16.5 & 19.9 & 20.0 & 0.0 & 37.4 & 11.2 \\
            \bottomrule
        \end{tabular}
    }
    \label{table:supp:rel_per_class_2dsg}
\end{table*}
    
\begin{table*}[b]
    \centering
    \caption{
        ORB-SLAM3 RMS ATE (cm) in each ReplicaSSG scene.
    }
    \vspace{-0.5em}
    \resizebox{\textwidth}{!}{
        \begin{tabular}{cccccc|c}
            \toprule
            \textbf{Apartment 0} & \textbf{Apartment 1} & \textbf{Apartment 2} & \textbf{Office 0} & \textbf{Office 1} & \textbf{Office 2} & \textbf{mean} \\
            4.6 & 1.9 & 3.8 & 0.9 & 0.5 & 4.0 & \multirow{5}{*}{3.6} \\[0.5em]
            \textbf{Office 3} & \textbf{Office 4} & \textbf{Room 0} & \textbf{Room 1} & \textbf{Room 2} & \textbf{Hotel 0} & \\
            3.1 & 1.9 & 0.9 & 0.9 & 1.3 & 2.9 & \\[0.5em]
            \textbf{FRL Apartment 0} & \textbf{FRL Apartment 1} & \textbf{FRL Apartment 2} & \textbf{FRL Apartment 3} & \textbf{FRL Apartment 4} & \textbf{FRL Apartment 5} & \\
            2.3 & 5.5 & 19.9 & 6.1 & 2.3 & 2.5 & \\
            \bottomrule
        \end{tabular}
    }
    \label{table:supp:orbslam_ate}
\end{table*}

\end{document}